\documentclass[lettersize, journal]{IEEEtran}
\usepackage{amsmath,amsfonts}
\usepackage{bm}
\usepackage[caption=false]{subfig}
\usepackage{multirow}
\usepackage[svgnames]{xcolor}
\usepackage{stfloats}
\usepackage{graphicx}
\usepackage{cite}
\usepackage{hyperref}
\usepackage[capitalize]{cleveref}
\usepackage{circledtext}
\usepackage{marvosym}
\begin{document}
\newcommand{\mycircledtext}[1]{\circledtext[charf=\large]{#1}}
\definecolor{mygreen}{RGB}{0, 180, 25}

\title{YOPO-Rally: A Sim-to-Real Single-Stage Planner for Off-Road Terrain}

\author{
    Hongyu Cao,
    Junjie Lu,
    Xuewei Zhang,
    Yulin Hui,
    Zhiyu Li,
    and Bailing Tian\textsuperscript{\Letter}
    \thanks{All authors are with the School of Electrical and Information Egineering, Tianjin University, Tianjin, 300072, China.}
}

\maketitle

\begin{abstract}

    Off-road navigation remains challenging for autonomous robots due to the harsh terrain and clustered obstacles.
    In this letter, we extend the YOPO (You Only Plan Once) end-to-end navigation framework to off-road environments, explicitly focusing on forest terrains, consisting of a high-performance, multi-sensor supported off-road simulator YOPO-Sim, a zero-shot transfer sim-to-real planner YOPO-Rally, and an MPC controller.
    Built on the Unity engine, the simulator can generate randomized forest environments and export depth images and point cloud maps for expert demonstrations, providing competitive performance with mainstream simulators.
    Terrain Traversability Analysis (TTA) processes cost maps, generating expert trajectories represented as non-uniform cubic Hermite curves.
    The planner integrates TTA and the pathfinding into a single neural network that inputs the depth image, current velocity, and the goal vector, and outputs multiple trajectory candidates with costs.
    The planner is trained by behavior cloning in the simulator and deployed directly into the real-world without fine-tuning.
    Finally, a series of simulated and real-world experiments is conducted to validate the performance of the proposed framework.
    The code of the simulator and planner will be released at \url{https://github.com/TJU-Aerial-Robotics/YOPO-Rally}.
    For supplementary video, please refer to \url{https://youtu.be/dyoufaKgVa0}.

\end{abstract}


\section{Introduction}

\IEEEPARstart{A}{utonomous} navigation has been widely applied to unmanned ground vehicles (UGVs) recently.
While the technology for indoor structured environments is mature, there are still many challenges for off-road unstructured environments, mainly for two reasons:
1) Terrain Traversability Analysis (TTA) relies on accurate sensors and is computationally intensive;
2) Complex terrain and obstacles affect the performance of the navigation system.
In addition, limited computational resources and low-accuracy sensors make it even more challenging.

Traditional off-road navigation frameworks mainly include three steps:
1) TTA to identify traversable areas and obstacles;
2) pathfinding to find a feasible and smooth path from the start to the goal;
3) a controller for the vehicle to follow the path \cite{OffRoadSurvey, OutdoorSurvey}.
Firstly, an accurate and efficient TTA is essential for the success of off-road navigation.
However, compared to high-accuracy long-range LiDAR, the noisy and short-range stereo camera significantly affects the accuracy and range of the cost map built from TTA\@.
Moreover, TTA is computationally intensive, which means it is slow and requires a lot of resources.
Secondly, the pathfinding, incorporating global and local planners, is employed to identify a feasible path from the start to the goal.
The global planner, which includes graph-based, sampling-based, and optimization-based search methods, is used to identify various reference paths based on the cost map.
This helps mitigate the local minima problem caused by the multi-modal nature of the navigation issue, where multiple equally valid paths may exist.
The local planner then follows these paths to generate a smooth trajectory.
Finally, the controller follows the reference path and executes the trajectory.
To reduce the TTA time, \cite{T-Hybrid} segments the raw point cloud map into an obstacle map and a terrain map, and only uses the terrain map for TTA, while \cite{PUTN, HD-RRT} performs TTA only on specific sampled points.
\cite{gp-navigation} constructs a Gaussian process-based local elevation map and performs TTA on it.
Though these modular frameworks are reliable and easy to debug \cite{NeuPAN}, they suffer from computational latency and error propagation.

\begin{figure}
  \centering
  \includegraphics[width=0.95\linewidth]{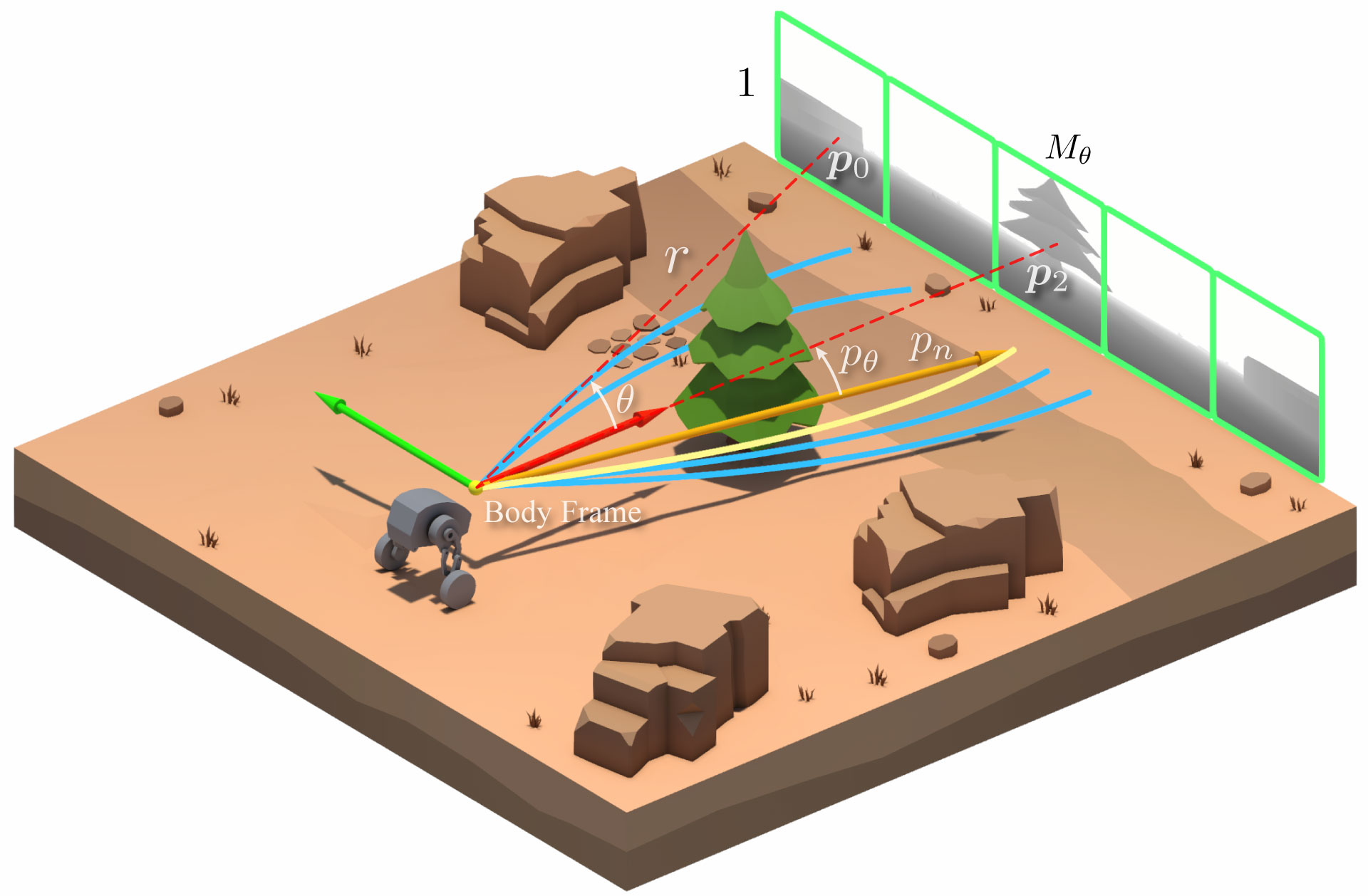}
  \caption{
    Visualization of the primitives and the predicted trajectories.
    The depth image projected to the far plane is divided into $M_\theta \times 1$ grids, corresponding to the primitive anchors $\bm{p}_i$ in the body frame.
    The end-state norm $p_n$ and angle $p_\theta$ predicted are the offsets to the primitive anchor $\bm{p}_i$.
  }
  \label{fig:primitive}
\end{figure}

With the rapid development of deep learning, end-to-end navigation frameworks that integrate multiple modules into a single network and map the sensor data directly to the trajectory alleviate these problems to some extent and show great potential.
Generally, it can be divided into two categories: 1) supervised learning and 2) reinforcement learning (RL).
It is known that deep learning requires extensive training data.
Supervised learning based methods learn from real-world or simulated data.
The former is expensive and time-consuming to collect, while the latter is easy to obtain but requires a high-fidelity simulator.
To alleviate the real-world data collection problem, \cite{BADGR} proposes a self-supervised learning method to collect real-world data automatically.
The RL-based methods generate data interactively in the simulator and learn the policy by trial and error.
Notably, end-to-end methods require a high-fidelity off-road simulator that provides multiple sensors.
However, mainstream simulators, i.e., CARLA \cite{CARLA}, and AirSim \cite{AirSim}, are designed for urban environments, and the off-road simulator is still in its infancy, due to the complexity and diversity of off-road environments.
In addition, the sim-to-real gap is another problem for simulated data training \cite{Sim2Real-Survey}.
For example, the image discrepancies stem from the illumination, texture, and noise differences between the real-world and simulated data, and the dynamics discrepancies stem from the model mismatch between the real-world and simulated vehicles.
It is worth noting that the sim-to-real gap is more severe for color images than for depth images, which are larger than for LiDARs.
Therefore, some end-to-end navigation methods choose depth image or LiDAR as input.
In \cite{WROOM}, a depth-based end-to-end navigation framework is proposed that uses RL to learn the policy from the simulated data and validates it in real-world environments.
In \cite{NeuPAN}, the navigation problem is formulated as an end-to-end MPC (Model Predictive Control) problem from the raw point cloud to the control commands, with a neural network as the solver.
To mitigate the dynamic discrepancies, \cite{AnycarToAnywhere} employs a neural network that inputs noisy historical states, noisy historical actions, and future actions, and predicts future states.
While \cite{AnyNav} focuses on the terrain-vehicle interaction, learning the friction coefficient from bird's eye view images.
Both are trained on simulated data, fine-tuned using real-world data, and demonstrate significant generalization ability.
Overall, the difficulty of collecting real-world data, the lack of high-fidelity off-road simulators, and the sim-to-real gap severely hamper the development of end-to-end navigation methods.

To address these challenges, we extend the YOPO (You Only Plan Once) \cite{YOPO} end-to-end navigation framework for off-road environments, explicitly focusing on forest terrains.
The YOPO framework, inspired by the YOLO (You Only Look Once) \cite{YOLO} object detection framework, integrates multiple modules\textemdash such as mapping, path searching, and trajectory optimization\textemdash into a single network, directly maps the sensor data to the offsets of the pre-defined motion primitive anchors to generate trajectories, like the bounding boxes in YOLO\@.
It consists of three components:
1) YOPO-Sim simulator, which provides a high-fidelity off-road forest environment for training and evaluation;
2) YOPO-Rally planner, which maps the depth image directly to the trajectory;
3) MPC controller, which executes the trajectory.
Firstly, the YOPO-Sim simulator generates randomized rugged terrains and trees using Perlin noise.
It collects the environmental point cloud map for each terrain and generates the depth image for each viewpoint sampled by Poisson disk sampling.
Secondly, the depth image is sent to the planner to reduce unnecessary color detail, possibly leading to a sim-to-real gap.
As visualized in \cref{fig:primitive}, the planner considers the terrain traversability and obstacles, and selects the best trajectory from the candidate trajectories.
Finally, the MPC controller executes the trajectory by optimizing the vehicle dynamics and kinematics.

In summary, the main contributions of this work are as follows:

\begin{enumerate}
  \item A high-fidelity, high-performance, and multi-sensor supported off-road simulator, YOPO-Sim, is proposed to simulate a vehicle driving in randomized off-road forest environments, with the capability of exporting the depth image and the environmental point cloud map.
  \item A learning-based single-stage planner, YOPO-Rally, is proposed to integrate Terrain Traversability Analysis (TTA) and pathfinding into a single network and generate multiple candidate trajectories considering the traversability and obstacles from noisy depth images.
  \item The planner learns from the simulated expert demonstrations and is transferred to the real-world without fine-tuning.
        A series of simulated and real-world experiments is conducted for validation.
\end{enumerate}

\section{Method}

\subsection{Single-Stage Planner}

\subsubsection{Trajectory Representation}
\label{sec:trajectory_representation}

This work represents the trajectory as a non-uniform cubic Hermite curve commonly used in computer graphics rather than the polynomial or the mapping matrix and selection matrix representation \cite{Gradient-based}.
With the Hermite curve, the trajectory $\bm{p}(t) = \left[p_x(t), p_y(t)\right]^{\mathrm{T}}$ is represented by a sequence of control points $\{\bm{p}_s, \bm{v}_s, \bm{v}_e, \bm{p}_e \}$ as:
\begin{subequations}
  \begin{align}
    \bm{p}(t) & = \mathbf{P} \mathbf{H} \mathbf{T} \label{eq:hermite_curve:position}            \\
              & = \begin{bmatrix}
                    \bm{p}_s & \bm{v}_s t_e & \bm{v}_e t_e & \bm{p}_e
                  \end{bmatrix}
    \begin{bmatrix}
      1 & 0 & -3 & 2  \\
      0 & 1 & -2 & 1  \\
      0 & 0 & -1 & 1  \\
      0 & 0 & 3  & -2
    \end{bmatrix}
    \begin{bmatrix}
      1      \\
      \tau   \\
      \tau^2 \\
      \tau^3
    \end{bmatrix} \, , \notag                                                                   \\
    \bm{v}(t) & =\mathbf{P} \mathbf{H} \mathbf{T}' / t_e \, , \label{eq:hermite_curve:velocity}
  \end{align}
\end{subequations}
where $\mathbf{P}$ is the control points matrix, $\mathbf{H}$ is the Hermite basis matrix, $\mathbf{T}$ is the time matrix, $\tau = t / t_e \in [0,1]$ is the normalized time, $\bm{p}_s = \left[p_s^x, p_s^x\right]^{\mathrm{T}}$ and $\bm{p}_e = \left[p_e^x, p_e^y\right]^{\mathrm{T}}$ are the start and end position of the trajectory, $\bm{v}_s = \left[v_s^x, v_s^y\right]^{\mathrm{T}}$ and $\bm{v}_e = \left[v_e^x, v_e^y\right]^{\mathrm{T}}$ are the start and end velocity, and $t_e$ is the pre-defined execution time of the trajectory.
Note that the Hermite curve representation naturally satisfies the boundary conditions of position and velocity: $\bm{p}(0) = \bm{p}_s$, $\bm{p}(t_e) = \bm{p}_e$, $\bm{v}(0) = \bm{v}_s$, $\bm{v}(t_e) = \bm{v}_e$.

\begin{figure}
  \centering
  \includegraphics[width=1\linewidth]{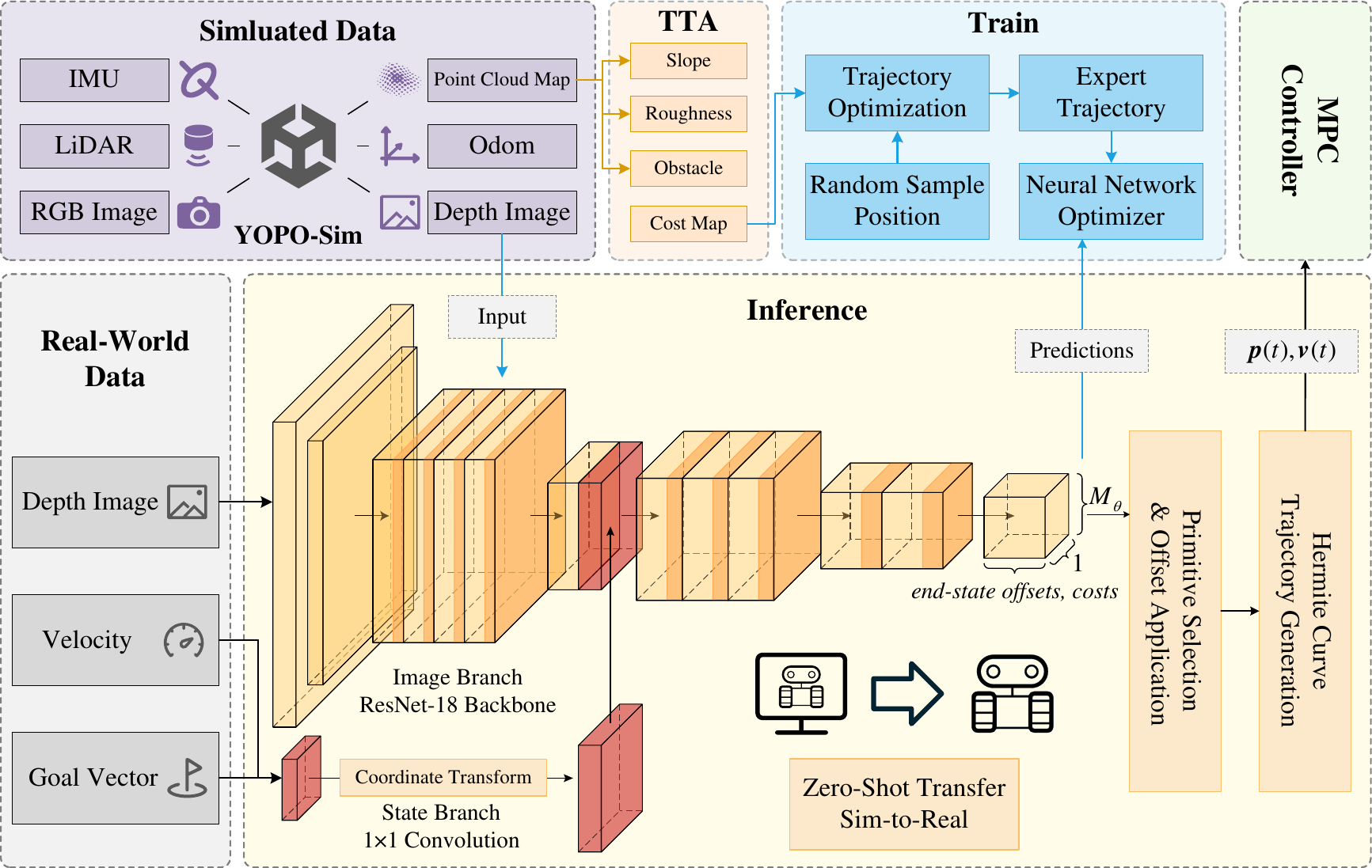}
  \caption{
    System overview.
    The planner inputs the depth image and the state consisting of velocity and goal vector, and predicts the end-state offsets and costs.
    Only the simulated data provided by YOPO-Sim is used for training, while the planner can be used in real-world scenarios without fine-tuning.
  }
  \label{fig:system_overview}
\end{figure}

The overall system is illustrated in \cref{fig:system_overview}.
As the navigation problem is in multi-modal, the planner may be trapped in local minima.
To address this issue, the traditional planner generates multiple trajectories with different initial guesses to cover the solution space.
Similarly, a set of motion primitives $P=\{\bm{p}_0, \bm{p}_1, \bm{p}_2, \dots, \bm{p}_{M-1}\}$ are selected as the initial guess anchors. The planner generates the offsets to each anchor, which generates multiple trajectories and selects the best one based on the cost.
Consistent with the previous work \cite{YOPO}, the primitive anchor is defined in the state lattice space.
Specifically, we uniformly sample $M_\theta$ angles within the horizontal field of view of the depth camera.
As visualized in \cref{fig:primitive}, the motion primitive anchor $\bm{p}_i$ is defined in the body frame as:
\begin{equation}
  \bm{p}_i = r[\cos \theta_i, \sin \theta_i]^{\mathrm{T}} \, ,
  \label{eq:primitive}
\end{equation}
where $i \in [0, M_\theta]$ is the index of the primitive, $r$ is the radius of the planning horizon.

Traditional planners usually require a cost map, which is difficult to obtain, to evaluate the safety and smoothness of the cost.
Then, a pathfinding algorithm minimizes the cost function, which requires multiple iterations.
In contrast, our planner is a single-stage planner that directly generates the trajectory without needing a pathfinding or a cost map.
For each primitive anchor $\bm{p}_i$, the planner predicts the end-state offsets $\Delta \bm{p}_e^i = \left(p_n^i, p_\theta^i\right)$ to the anchor $\bm{p}_i$ as well as the end-state velocity $\bm{v}_e^i$ and its cost $c_i$.
The end-state position $\bm{p}_e^i$ is calculated as:
\begin{equation}
  \bm{p}_e^i =p_n^i \left[\cos \left(\theta_i + p_\theta^i\right), \sin \left(\theta_i + p_\theta^i\right)\right]^{\mathrm{T}} \, .
  \label{eq:end_state}
\end{equation}

Then the planner selects the best primitive anchor based on the cost and generates the trajectory along with the start-state position $\bm{p}_s$ and velocity $\bm{v}_s$ following the Hermite curve representation in \cref{eq:hermite_curve:position,eq:hermite_curve:velocity}.

\subsubsection{Network Architecture}
\label{sec:network_architecture}

In this work, multiple procedures are integrated into a single neural network to predict the trajectory directly.
As shown in \cref{fig:system_overview}, the network consists of the feature extractor and the trajectory generator.
It inputs depth image, current velocity, and goal direction vector, while outputting the end-state offsets, velocity, and the cost, which are represented in the body frame to reduce the dimensionality.
Firstly, for the image branch, the depth image is divided into $M_\theta \times 1$ lattices, with the primitive anchors in the center of each lattice.
The network predicts the offsets $p_n, p_\theta$, the velocity $\bm{v}_e$, and the cost $c$ for each anchor.
The modified ResNet-18 feature extractor inputs the depth image and outputs the feature map.
Specifically, the last two layers of the ResNet-18 are removed, and the channel number of the first layer is modified to $1$ with the depth image resolution of $32 M_\theta \times 32$ and the feature map resolution of $M_\theta \times 1$.
Secondly, for the state branch, the state consisting of the current velocity $\bm{v}_s$ and goal direction vector $\bm{g}$ is extracted with a $1 \times 1$ convolutional layer to form the state feature map and concatenated with the feature map.
Thirdly, for the trajectory generator, $3$ convolutional layers with the last layer having different activation functions are used to predict the end-state offsets, velocity, and cost.
As the end-state offset norm meets $p_n \in [0, p_{n_{max}}]$ and cost meets $c \in [0, c_{max}]$, the sigmoid function is used as the activation function. In contrast, the end-state offset angle meets $p_\theta \in [-p_{\theta_{max}}, p_{\theta_{max}}]$ and velocity $\bm{v}_e = \left[v_x, v_y\right]^{\mathrm{T}}$ meet $v_x, v_y \in [-v_{max}, v_{max}]$, the tanh function is used as the activation function.

To enhance the performance, the input image, input state $\bm{s}$, and output $\bm{y}$ are normalized as:
\begin{equation}
  \bm{s}  = \begin{bmatrix}
    \bm{v}_s / v_{max} \\
    \bm{g} / g_{max}   \\
  \end{bmatrix} \, , \quad
  \bm{y}  = \begin{bmatrix}
    p_n / p_{n_{max}}           \\
    p_\theta / p_{\theta_{max}} \\
    \bm{v}_e / v_{max}          \\
    c / c_{max}                 \\
  \end{bmatrix} \, .
  \label{eq:normalize}
\end{equation}

To generate the trajectory, the primitive anchor with the minimum cost is selected as the final output, and the end-state offsets $\bm{p}_e$ are restored according to \cref{eq:end_state} and then transformed into the world frame.
The trajectory $\bm{p}(t)$ is calculated by \cref{eq:hermite_curve:position}.

\subsection{Training Strategy}

Behavior cloning (BC) is used to train the single-stage planner, which learns from expert demonstrations.
Since there is a little discrepancy between the simulated and real-world depth images, it is enough to train the planner in the simulator and directly apply it to the real-world without fine-tuning, which is zero-shot sim-to-real transfer.

To acquire the observation data, YOPO-Sim simulates the vehicle in randomized off-road forest environments.
The depth image, position, and orientation of the vehicle are recorded at each viewpoint, and the point cloud map of each environment is recorded as well.
To augment the data, the input state $\bm{s}$ is randomly sampled several times at each viewpoint.

To train the policy, a trajectory optimization algorithm is used to generate a set of expert trajectories.
Specifically, Terrain Traversability Analysis (TTA) is performed on the point cloud map to rasterize it and generate a 2D cost map, then the objective function and optimization variables are defined. Finally, the optimization problem is solved to create the expert trajectories.

The geometry feature (slope $G_s$ and roughness $G_r$) of the terrain point cloud map is processed with CloudCompare \cite{CloudCompare} for TTA, followed by rasterizing the result to generate the DEM (Digital Elevation Map), consisting of the elevation map, slope map, and roughness map.

The overflow slope and roughness then identify the obstacle map $V_c$ as:
\begin{equation}
  V_c(\bm{p}) = \begin{cases}
    1, & G_s(\bm{p}) > G_{s_{max}} \text{ or } G_r(\bm{p}) > G_{r_{max}} \, ; \\
    0, & \text{otherwise} \, ,
  \end{cases}
\end{equation}
where $G_{s_{max}}$ and $G_{r_{max}}$ are the threshold values for the slope and roughness, and $\bm{p}$ is the point in the DEM\@.
Then the obstacle map is dilated by a circular kernel with radius $r_{dilate}$ to ensure the safety margin around the obstacle, followed by the SDF (Signed Distance Field) calculation for the distance $D_t$ to the obstacle.
The safety margin $d_0$ and the decay rate $k$ are used to calculate the safety cost $C_{safety}$ as:
\begin{equation}
  C_{safety} = e^{(d_0 - D_t) / k} \, .
  \label{eq:safety_cost}
\end{equation}

The cost map $C_{map}$ is generated by combining the geometry cost and safety cost with the following formula:
\begin{equation}
  C_{map}  =  \lambda_r \frac{G_r}{G_{r_{max}}} + \lambda_s \frac{G_s}{G_{s_{max}}} + \lambda_{c} C_{safety} \, ,
  \label{eq:cost_map}
\end{equation}
where $\lambda_r, \lambda_s, \lambda_{c}$ are the roughness, slope, and safety cost weights, respectively.

The bicubic interpolation instead of the bilinear interpolation is used to generate a continuous cost map $\widetilde{C}_{map}$ because the former is differentiable everywhere while the latter is not differentiable at the interpolation nodes, which is detrimental to gradient-based optimizers and may even lead to optimization failure.

The objective function $J_t$ is defined as the sum of the map cost and the control cost along the trajectory, and the terminal cost to the goal position:
\begin{equation}
  \begin{gathered}
    J_t(\bm{p}_e, \bm{v}_e)  = \int_{0}^{t_e} \widetilde{C}_{map}(\bm{p}(t)) + \left\lVert \bm{v}(t)\right\rVert ^2 \, \mathrm{d}t
    + \left\lVert \bm{p}_e - \bm{g}\right\rVert ^2 \\
    \approx \sum_{n = 0}^{N_t} \left[\widetilde{C}_{map}(\bm{p}(t_n)) + \left\lVert \bm{v}(t_n)\right\rVert ^2 \right] \Delta T
    + \left\lVert \bm{p}_e - \bm{g}\right\rVert ^2 \, ,
  \end{gathered}
  \label{eq:objective_function}
\end{equation}
where $N_t$ is the number of time steps, $\Delta T = t_e / N_t$ is the time step, $t_n = t_e \cdot n  / N_t$ is the time at step $n$, $\bm{p}(t), \bm{v}(t)$ are the trajectory and the velocity calculated by \cref{eq:hermite_curve:position,eq:hermite_curve:velocity}.

\begin{figure}
  \centering
  \includegraphics[width=0.95\linewidth]{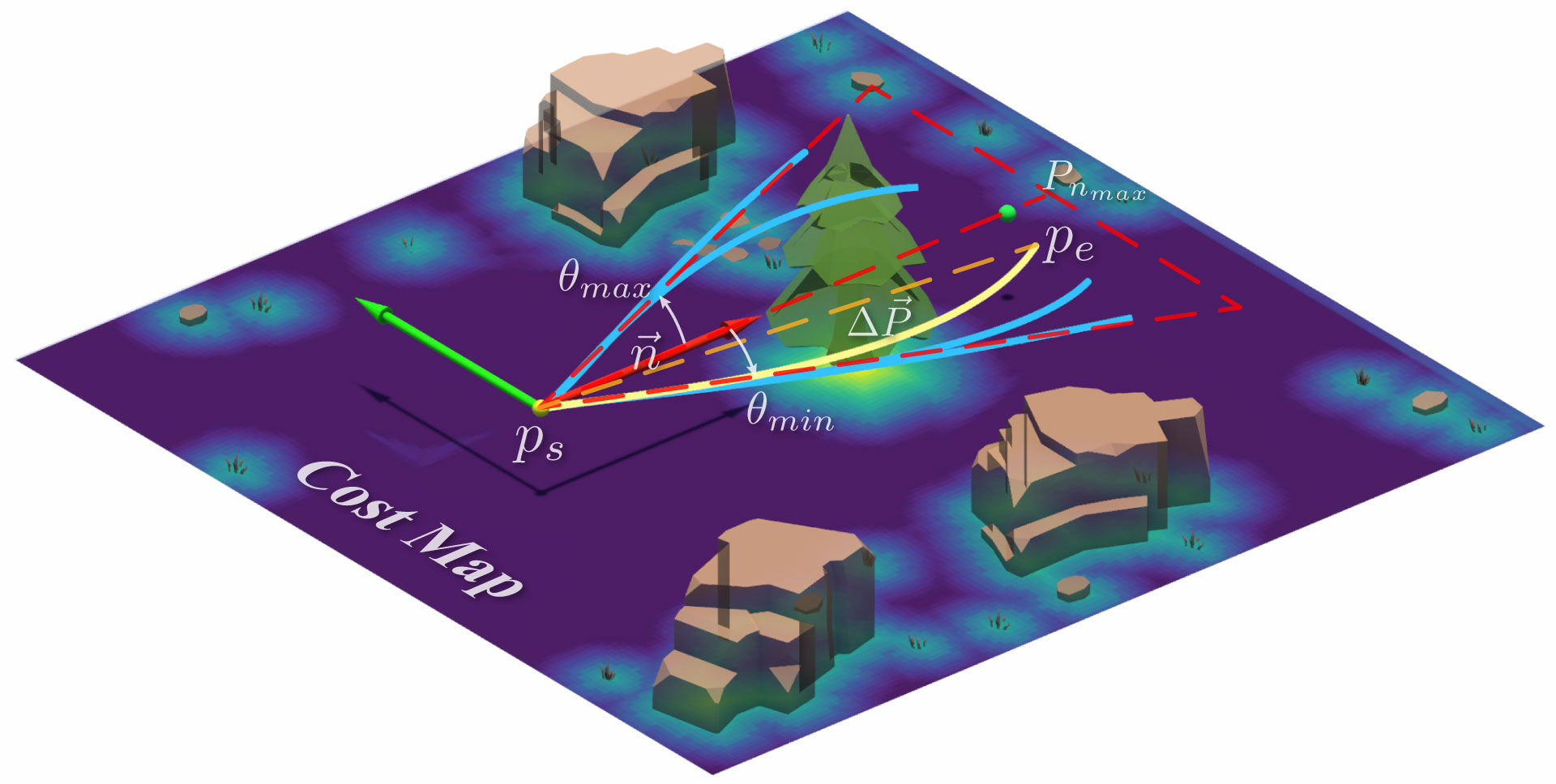}
  \caption{
    Visualization of the cone constraint, cost map, and expert trajectories.
    The cone constraint is defined by the primitive anchor norm $P_{n_{max}}$ and the direction vector $\bm{n}$.
    The offset $\Delta \bm{P} $ is constrained to the cone's interior.
    The cost map is generated by TTA, which evaluates the cost function in \cref{eq:objective_function}.
    The trajectory optimization problem is solved by CasADi \cite{CasADi}.
    For a better illustration, please refer to the video.
  }
  \label{fig:trajectory_optimization}
\end{figure}

As can be seen from the neural network architecture in \cref{sec:network_architecture}, the end position $\bm{p}_e$ of the output $M_\theta \times 1$ trajectories lies inside the cone composed of the primitive anchor with the apex at the start position $\bm{p}_s$, the center of the circular base at the primitive anchor, the half angle $\theta_{half}$, the height length $P_{n_{max}}$ and the height vector $\bm{n}$.
To match the neural network's output range, each trajectory to be optimized is constrained to the cone's interior, thus generating the expert demonstration for each trajectory and avoiding label assignment.
As shown in \cref{fig:trajectory_optimization}, the cone constraint is inspired by the gaze target constraint in Drake \cite{drake} and is formulated as:
\begin{equation}
  \begin{gathered}
    0 < \Delta \bm{P} \cdot \bm{n} < P_{n_{max}} \, ,                                                                     \\
    \cos ^2 \theta_{half} \left\lVert \Delta \bm{P}\right\rVert ^2 \leq \left(\Delta \bm{P} \cdot \bm{n}\right) ^2 \, ,
  \end{gathered}
  \label{eq:cone_constraint}
\end{equation}
where $\Delta \bm{P} = \bm{p}_e - \bm{p}_s$ is the offset from the start position to the end position, $\bm{n} = (\cos \theta_{center}, \sin \theta_{center})$ is the normalized height vector, $\theta_{max}$ and $\theta_{min}$ are the angles of the two generatrixes of the cone intersecting the 2D planning plane, $\theta_{center} = ( \theta_{max} + \theta_{min} ) / 2$ is the center angle of the primitive anchor, and $\theta_{half} = ( \theta_{max} - \theta_{min} ) / 2$ is the half angle of the primitive anchor.
The trajectory optimization problem is formulated as:
\begin{subequations}
  \begin{align}
    \min \limits_{\bm{p}_e, \bm{v}_e} \quad & J_t(\bm{p}_e, \bm{v}_e) \, ,                                                                        \\
    \mathrm{s.t.} \quad                     & \left\lVert \bm{v}_e\right\rVert ^2 \leq v_{max}^2 \, , \label{eq:trajectory_optimization:velocity} \\
                                            & 0 < \Delta \bm{P} \cdot \bm{n} < P_{n_{max}} \, , \label{eq:trajectory_optimization:cone_norm}      \\
                                            & \cos ^2 \theta_{half} \left\lVert \Delta \bm{P}\right\rVert ^2 \leq
    \left(\Delta \bm{P} \cdot \bm{n}\right) ^2 \, , \label{eq:trajectory_optimization:cone_angle}
  \end{align}
  \label{eq:trajectory_optimization}
\end{subequations}
where \cref{eq:trajectory_optimization:velocity} is the velocity constraint, \cref{eq:trajectory_optimization:cone_norm,eq:trajectory_optimization:cone_angle} together form the cone constraint.
This problem is solved by CasADi \cite{CasADi} to generate the expert trajectories for training.

The loss function is defined as the mean squared error between the predicted trajectory $\bm{y}_n$ and the expert trajectory $\bm{y}_n^*$:
\begin{equation}
  \mathcal{L} = \frac{1}{N} \sum_{n = 0}^{N} \left\lVert \bm{y}_n - \bm{y}_n^*\right\rVert ^2 \, ,
  \label{eq:loss_function}
\end{equation}
and the AdamW \cite{AdamW} optimizer is used to minimize the loss function.

\subsection{MPC Controller}

The Model Predictive Control (MPC) based controller is used to execute the trajectory generated by the single-stage planner.
The trajectory following problem is formulated as an MPC problem, where the objective is to minimize the deviation from the desired trajectory and the control effort, while satisfying the constraints of the vehicle dynamics and kinematics constraints.

The differential-drive vehicle model is used to describe the vehicle dynamics $\dot{\bm{x}}=\bm{f}(\bm{x},\bm{u})$, where $\bm{x} = \left[x, y, \theta \right]^\mathrm{T}$ is the state vector and $\bm{u} = \left[ v, w \right]^\mathrm{T}$ is the control input vector. The state vector represents the vehicle's position $(x,y)$ and orientation $\theta$, while the control input vector represents the linear velocity $v$ and angular velocity $w$. The following equation can describe the vehicle dynamics:
\begin{equation}
    \bm{f}(\bm{x},\bm{u}) = \begin{bmatrix}
        \cos\theta & 0 \\
        \sin\theta & 0 \\
        0          & 1
    \end{bmatrix}
    \begin{bmatrix}
        v \\
        w
    \end{bmatrix} \, .
    \label{eq:vehicle_dynamics}
\end{equation}

The MPC problem is formulated as:
\begin{subequations}
    \begin{align}
        \min \limits_{\bm{x}, \bm{u}} \quad & \sum_{k=0}^{N - 1}\left( \left\lVert \bm{x}_k - \bm{x}_k^r\right\rVert _{\mathbf{Q}} + \left\lVert \bm{u}_k\right\rVert _{\mathbf{R}} \right) \, , \\
        \mathrm{s.t.} \quad                 & \bm{x}_{k+1} = \bm{x}_k + \Delta t \cdot \bm{f}(\bm{x}_k, \bm{u}_k) \, ,                                                                           \\
                                            & \bm{x}_0 = \bm{x}_{init} \, ,                                                                                                                      \\
                                            & \bm{u}_{k} \in \mathcal{U} \, ,
    \end{align}
\end{subequations}
where $\bm{x}_{init}$ is the initial state and $\Delta t$ is the time step. The matrices $\mathbf{Q}$ and $\mathbf{R}$ are weighting matrices.
The optimization problem is solved by CasADi \cite{CasADi} to find the optimal control inputs $\bm{u}^*$, which are then applied to the vehicle to follow the desired trajectory.

\subsection{Simulation Environment}

\begin{figure*}
  \centering
  \subfloat[Depth Camera and RGB Camera]{
    \includegraphics[width=0.3\linewidth]{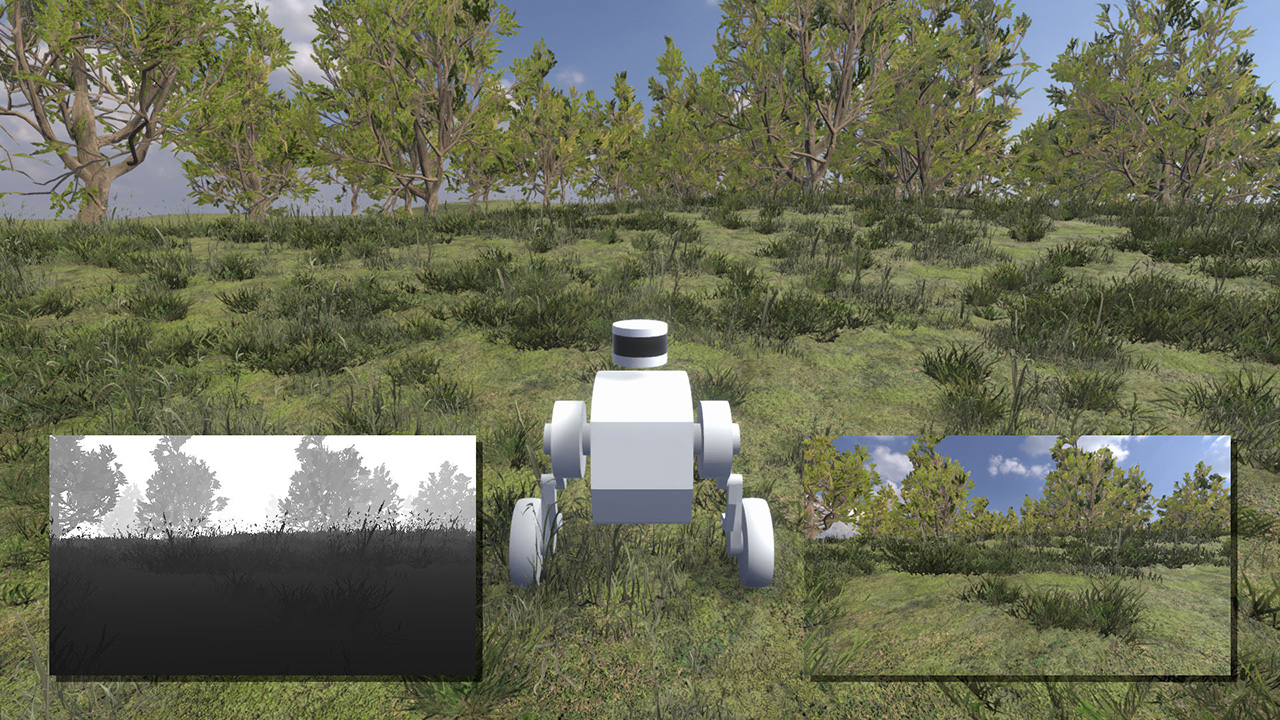}
    \label{fig:simulation_environment:sensors}
  }
  \hfil
  \subfloat[LiDAR]{
    \includegraphics[width=0.3\linewidth]{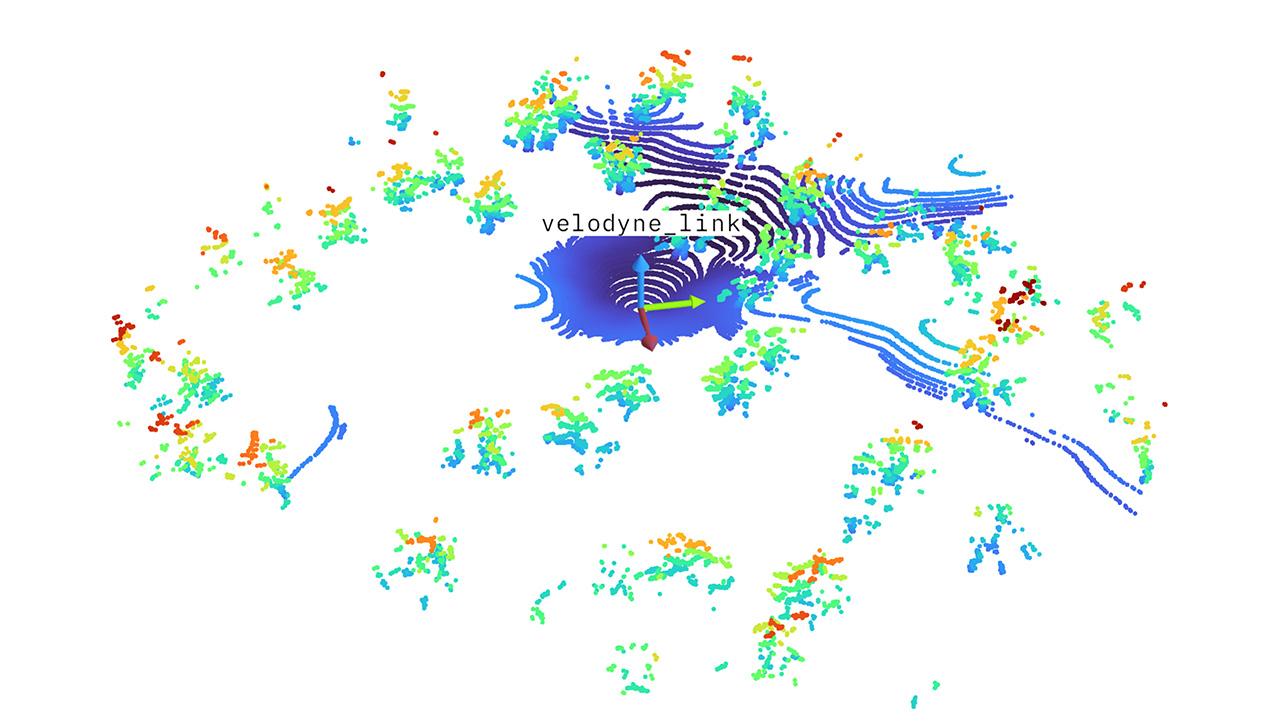}
    \label{fig:simulation_environment:lidar}
  }
  \subfloat[Point Cloud Map]{
    \includegraphics[width=0.3\linewidth]{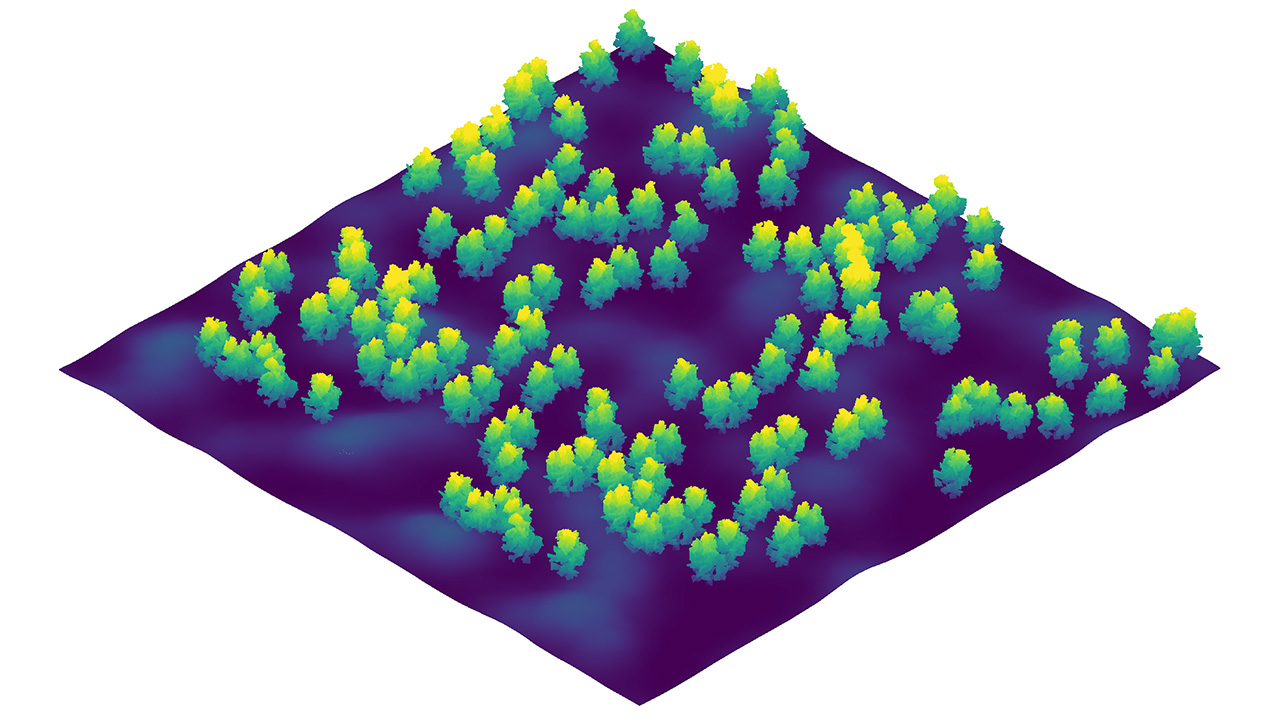}
    \label{fig:simulation_environment:voxel_generator}
  }
  \caption{
    Snapshots of the YOPO-Sim.
    The insets show some of the sensors provided by the simulator: (a) depth camera and RGB camera, (b) LiDAR, and (c) point cloud map.
  }
  \label{fig:simulation_environment}
\end{figure*}

The YOPO-Sim simulator is built upon the Unity 3D game engine, as shown in \cref{fig:simulation_environment}, and is designed to provide a realistic simulation environment for off-road forest driving tasks.
It simulates a vehicle with various sensors (e.g., depth camera, RGB camera, LiDAR, and IMU) driving in randomized off-road forest environments, and generates the point cloud map.
The ROS-TCP-Connector package is used to connect the simulator with ROS, allowing for seamless communication between the simulator and the ROS ecosystem.

The sensors are provided by the UnitySensors package \cite{UnitySensors}, an open-source Unity plugin that offers various high-performance sensors for robotics research.
This work uses the depth camera to generate the depth image for training.

The voxel generator generates the point cloud map, which is essential for TTA.
Inspired by Flightmare \cite{flightmare}, the voxel generator is implemented efficiently using Unity's job system and the Burst compiler.
In short, the voxel generator rasterizes the space to generate the voxel grid and traverses each voxel to check if it is occupied.
The occupied voxel is appended to the point cloud map.
The 2D and 3D voxel generators are provided, and the former is suitable for 2D terrain environments, while the latter is suitable for 3D space environments.
For a terrain of size $L \times W \times H$ m$^3$, the voxel grid to traverse is generated with the number of $N=(L \times W \times H) / l^3$, where $l$ is the voxel size.
Though the computational complexity is $\mathcal{O}(n^2)$ for the 2D voxel generator and $\mathcal{O}(n^3)$ for the 3D voxel generator, where $n$ is the side length of the voxel grid, it is still efficient enough to generate the point cloud map in near real time in parallel.
In this work, the 2D voxel generator generates the terrain point cloud map, and the 3D voxel generator creates the obstacle point cloud map for TTA.

To create randomized off-road forest environments, the terrain and trees are generated by the Perlin noise algorithm provided by the Vista package.
The Poisson disk sampling algorithm is used to get uniformly distributed sampling points.

\section{Experiment}

This section conducts a series of experiments to evaluate the performance of the proposed off-road simulator, YOPO-Sim, and learning-based planner, YOPO-Rally.
The planner runs at 10 Hz, and the MPC controller at 20 Hz.
The simulator and simulated experiments are conducted on a PC with an Intel i7-13700KF CPU and an NVIDIA RTX 4070 GPU\@.
As shown in \cref{fig:diablo_robot}, the real-world experiments are conducted on the DIABLO robot equipped with an OAK-D-Pro stereo camera for depth image, a T265 camera for odometry, and an NVIDIA Jetson Orin NX for computing.
The snapshots of the simulated and real-world experiments are shown in \cref{fig:experiment:snapshots}.

\subsection{Simulator Evaluation}

\begin{table*}
    \tabcolsep=0.3cm
    \centering
    \caption{Sensor Performance}
    \label{tab:sensor_performance}
    \resizebox{0.93\linewidth}{!}{
        \begin{tabular}{|c|c|c|c|c|c|c|c|c|c|c|}
            \hline
            \multicolumn{2}{|c|}{Sensor}                                           & \multicolumn{6}{c|}{Camera
            (resolution/scan)}                                                     & \multicolumn{3}{c|}{LiDAR
            (points/scan)}                                                                                                                                                                                                                                                                                                       \\
            \hline
            \multicolumn{2}{|c|}{\multirow{2}{*}{Configuration}}                   & \multicolumn{3}{c|}{Depth
            Camera}                                                                & \multicolumn{3}{c|}{RGB
            Camera}                                                                & Mid
            360                                                                    & VLP-16                     & VLP-32                                                                                                                                                                                                         \\
            \cline{3-11}
            \multicolumn{2}{|c|}{}                                                 & 360p                       & 720p                 & 1080p                & 360p                 & 720p                 & 1080p                & 20K                  & 57.6K                 & 115.2K                                       \\
            \hline
            \multirow{4}{*}{\begin{tabular}[c]{@{}c@{}}Update\\ Rate\end{tabular}} & AirSim                     & 37                   & 14                   & 7                    & 74                   & 46                   & 27                   & 22                    & 9                     & 5                    \\
            \cline{2-11}
                                                                                   & Flightmare                 & 24                   & 13                   & 9                    & 44                   & 23                   & 21                   & /                     & /                     & /                    \\
            \cline{2-11}
                                                                                   & CARLA                      & \textcolor{red}{140} & \textcolor{red}{134} & \textcolor{red}{57}  & \textcolor{red}{117} & \textcolor{red}{102} & \textcolor{red}{70}  & \textcolor{blue}{182} & \textcolor{red}{157}  & \textcolor{blue}{97} \\
            \cline{2-11}
                                                                                   & Proposed                   & \textcolor{blue}{95} & \textcolor{blue}{64} & \textcolor{blue}{52} & \textcolor{blue}{78} & \textcolor{blue}{65} & \textcolor{blue}{48} & \textcolor{red}{380}  & \textcolor{blue}{145} & \textcolor{red}{122} \\
            \hline
        \end{tabular}
    }
\end{table*}

\begin{table}
    \centering
    \renewcommand\arraystretch{1.2}
    \caption{Voxelization Performance}
    \label{tab:voxelization_performance}
    \resizebox{0.85\linewidth}{!}{
        \begin{tabular}{|c|c|c|c|c|c|}
            \hline
            \multicolumn{2}{|c|}{Voxel size (m)}     & 0.3        & 0.2   & 0.15   & 0.12           \\
            \hline
            \multicolumn{2}{|c|}{Total Voxel Number} & 16.7M      & 56.3M & 133.3M & 260.4M         \\
            \hline
            \multicolumn{2}{|c|}{Point Cloud Length} & 0.7M       & 2.2M  & 5.2M   & 10.0M          \\
            \hline
            \multirow{2}{*}{Time (s)}                & Flightmare & 7.01  & 20.47  & 45.23  & 85.31 \\
            \cline{2-6}
                                                     & Proposed   & 0.35  & 1.19   & 2.83   & 6.03  \\
            \hline
        \end{tabular}
    }
\end{table}

\subsubsection{Sensor Performance}

In this section, the performance of the sensor component of YOPO-Sim is evaluated, including the depth camera, RGB camera and LiDAR sensors.
A comparison has been made with some popular simulators, including AirSim \cite{AirSim}, Flightmare \cite{flightmare}, and CARLA \cite{CARLA} at similar workloads.
The maximum \emph{sensor update rate}\textemdash not the \emph{render frame rate}, since the simulator does not acquire sensor data in every render frame but only when triggered, nor the \emph{physics update rate}, as the physics engine is not the bottleneck in the simulation\textemdash is measured for different configurations, including various resolutions for the depth camera and RGB camera, and other numbers of scan points for the LiDAR, as shown in \cref{tab:sensor_performance}.
The best performance is highlighted in red and the second best performance is highlighted in blue.
The experiment lasted for 10 seconds and the average update rate is reported.
The results show that the proposed simulator achieves a better performance than AirSim and Flightmare, and is comparable to CARLA.

For the depth camera and RGB camera, the proposed simulator and CARLA achieve the best performance.
This is because the proposed simulator and CARLA use asynchronous texture requests, which improve the performance, while AirSim and Flightmare use synchronous ones.
In addition, AirSim and Flightmare fall short in depth cameras compared to RGB cameras.
This is because they encode the depth image pixels as 32-bit float values, which is not efficient for transmission, whereas the proposed simulator encodes them as 16-bit unsigned integers, and CARLA encodes them as three 8-bit unsigned integers.

As for the LiDAR sensors, the proposed simulator and CARLA achieve significantly better performance than AirSim, which is because parallel ray cast is used in the proposed simulator and CARLA, while a single-threaded for-loop ray cast is used in AirSim.

\subsubsection{Voxelization Performance}

Since the point cloud map generated by the voxel generator is crucial for training the neural network, it is important to evaluate the performance.
It is compared with the voxel generator in Flightmare.
The simulated environment is a randomly generated forest with a size of $150 \times 150 \times 20$ m$^3$, as shown in \cref{fig:simulation_environment:voxel_generator}.
The time taken to generate a voxel point cloud map with different voxel sizes is measured in the \cref{tab:voxelization_performance}.
The results are tested 10 times, and the average execution time is reported, showing that our voxel generator is significantly faster than the one in Flightmare.

This is because the voxel generator in Flightmare uses a single-threaded for-loop, which is not efficient for large-scale voxelization. In contrast, the proposed voxel generator is parallelized by Unity's job system and highly optimized by the Burst compiler.

\begin{figure}
    \centering
    \includegraphics[width=0.85\linewidth]{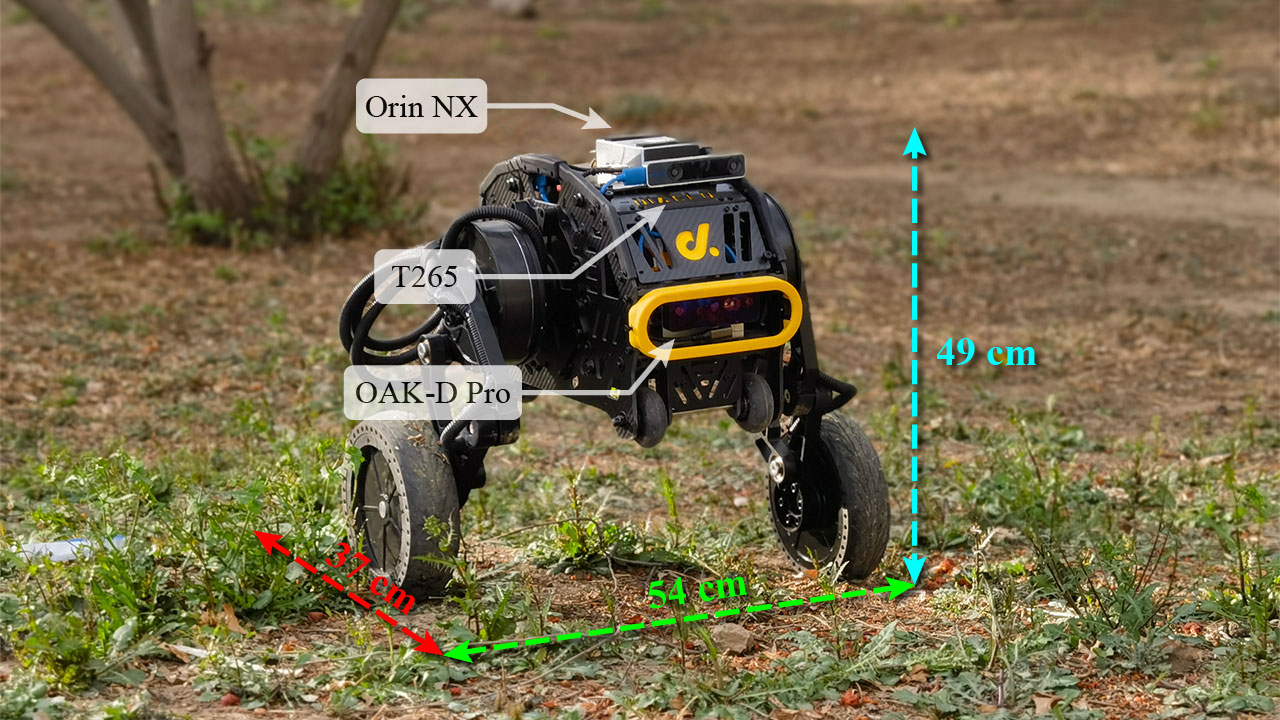}
    \caption{
        The DIABLO vehicle platform for the real-world experiment is equipped with a RealSense T265 camera for odometry and an OAK-D-Pro stereo camera for depth image.
    }
    \label{fig:diablo_robot}
\end{figure}

\begin{figure*}
    \centering
    \subfloat[Simulated view in \cref{fig:simulated_comparison:dense} \mycircledtext{1}.]{
        \begin{minipage}{0.3\linewidth}
            \includegraphics[width=\linewidth]{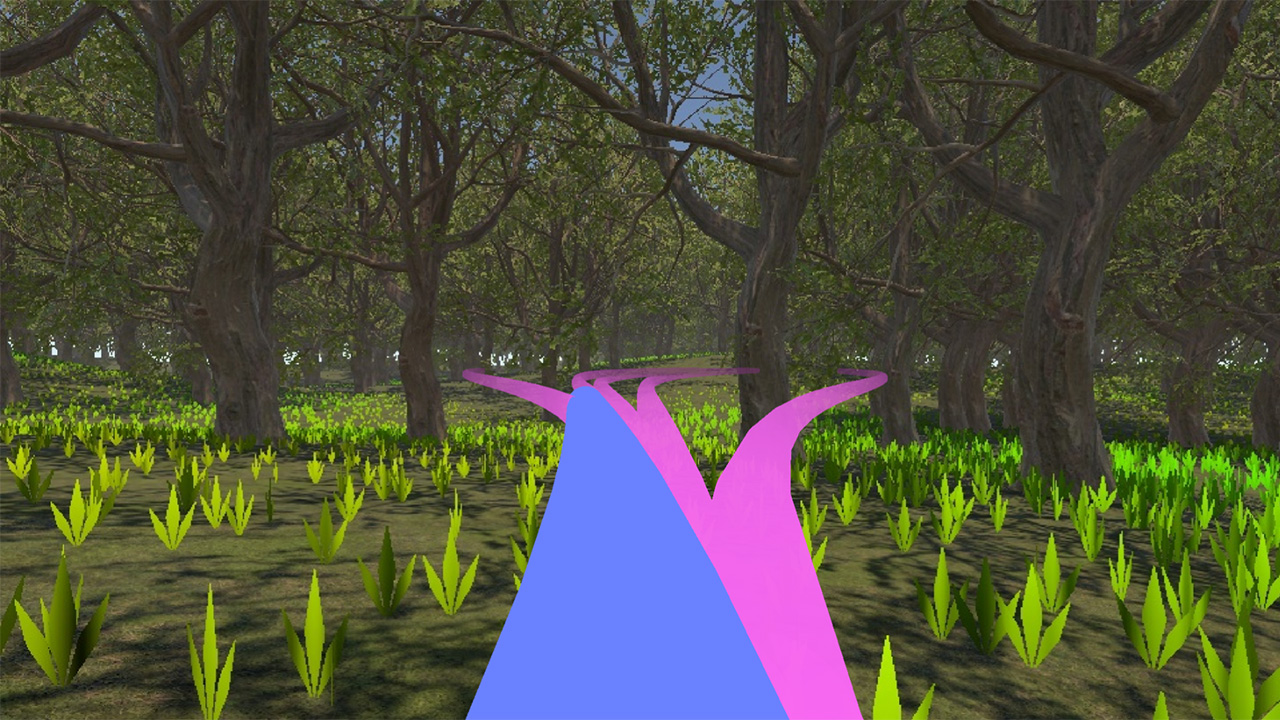}%
            \vspace{2pt}
            \includegraphics[width=\linewidth]{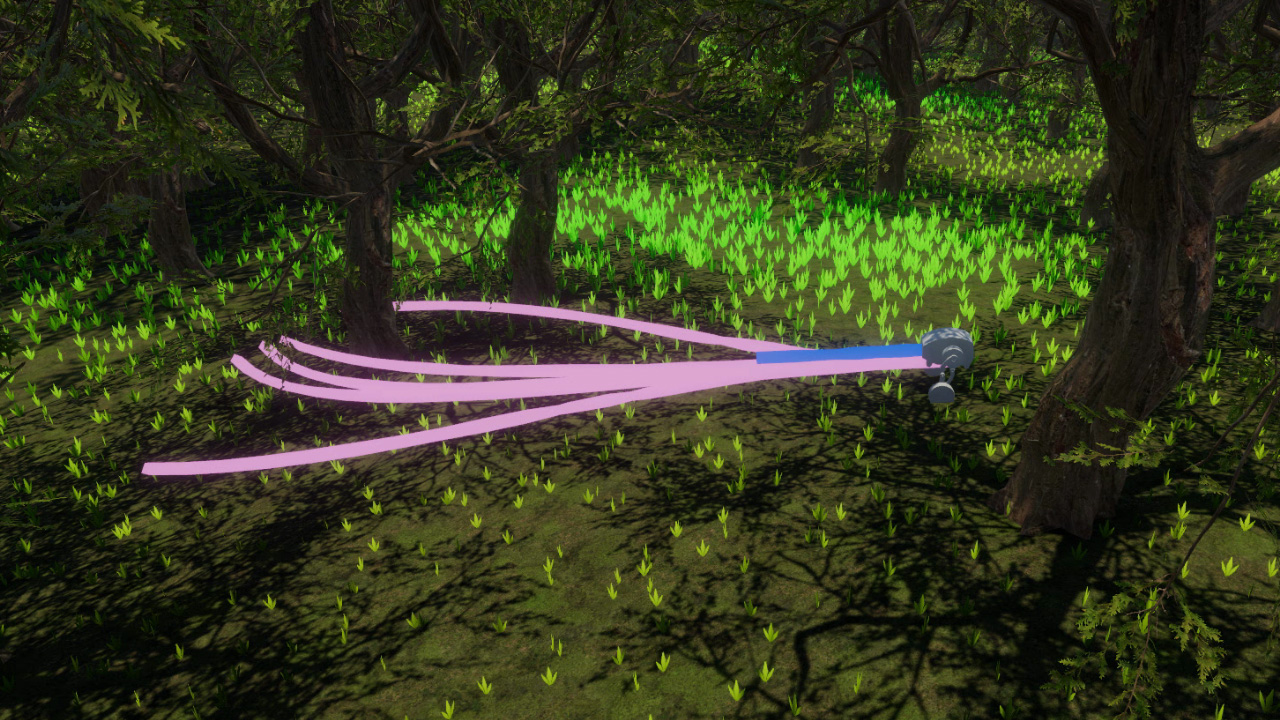}
        \end{minipage}
        \label{fig:experiment:snapshots:sparse}
    }\hspace{-8pt}
    \subfloat[Real-world view in \cref{fig:real_world_experiment:bag11} \mycircledtext{2}.]{
        \begin{minipage}{0.3\linewidth}
            \includegraphics[width=\linewidth]{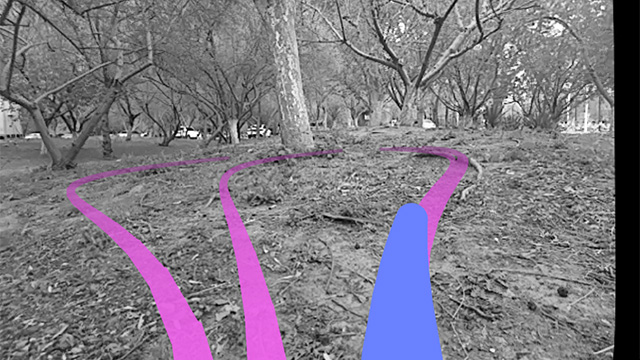}%
            \vspace{2pt}
            \includegraphics[width=\linewidth]{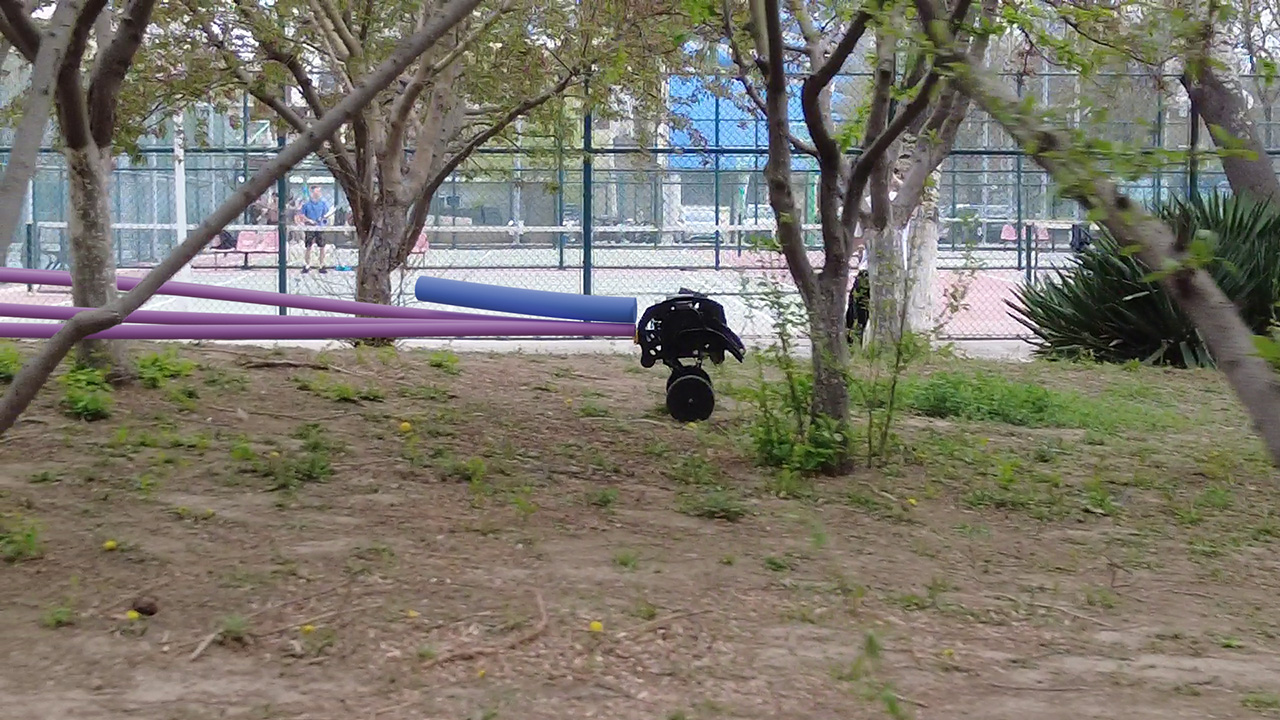}
        \end{minipage}
        \label{fig:experiment:snapshots:bag11}
    }\hspace{-8pt}
    \subfloat[Real-world view in \cref{fig:real_world_experiment:bag9} \mycircledtext{3}.]{
        \begin{minipage}{0.3\linewidth}
            \includegraphics[width=\linewidth]{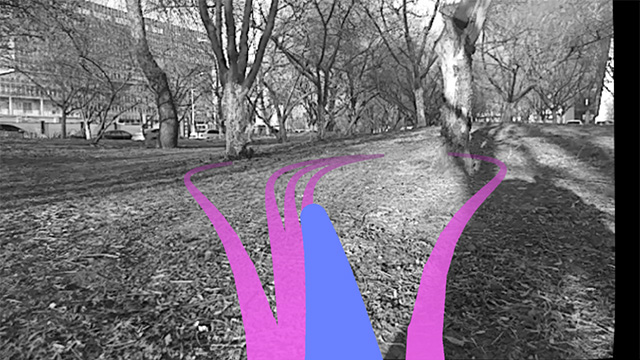}%
            \vspace{2pt}
            \includegraphics[width=\linewidth]{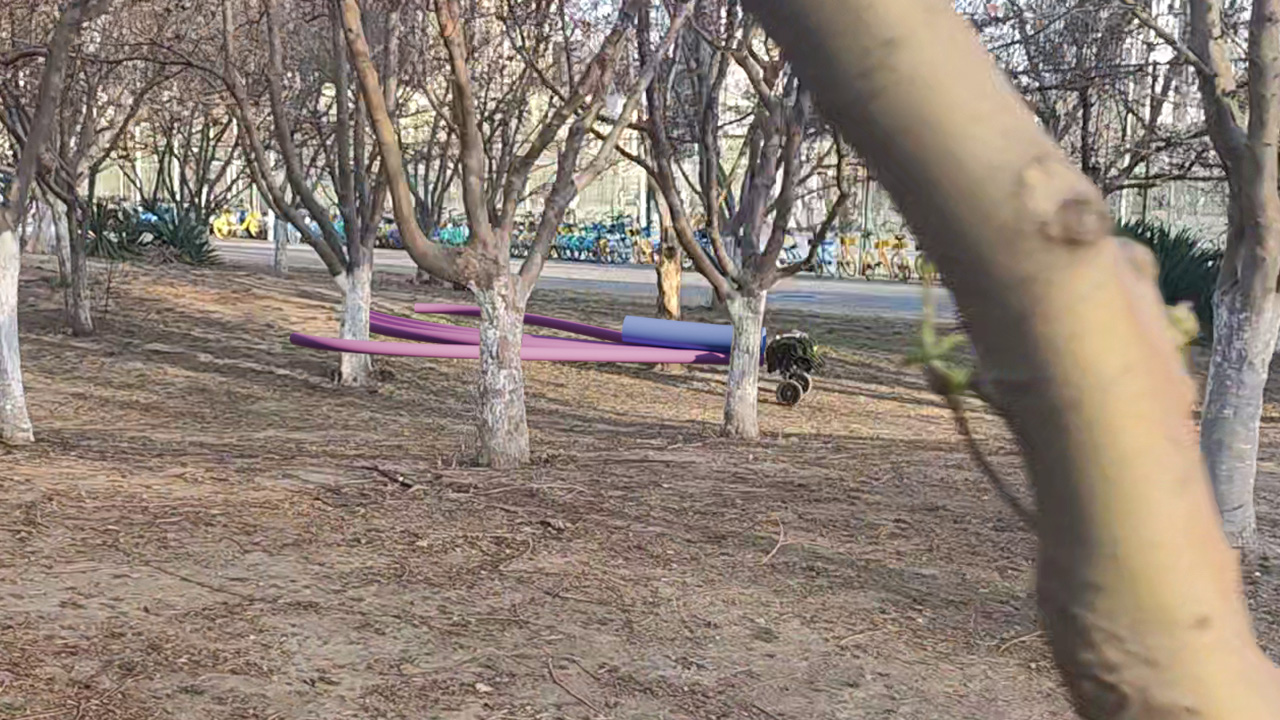}
        \end{minipage}
        \label{fig:experiment:snapshots:bag9}
    }
    \caption{
        The simulated and real-world experimental views of the robot in different environments.
        \mycircledtext{1}, \mycircledtext{2}, and \mycircledtext{3} are the snapshots of the corresponding positions in \cref{fig:simulated_comparison:dense,fig:real_world_experiment:bag11,fig:real_world_experiment:bag9}, respectively.
        The trajectories are overlaid on the images, where the purple curves are the planner's candidate trajectories and the blue curves are the MPC predicted trajectories.
    }
    \label{fig:experiment:snapshots}
\end{figure*}

\subsection{Simulated Comparison}

\begin{table*}
    \renewcommand\arraystretch{1.1}
    \centering
    \caption{Simulated Comparison}
    \label{tab:simulated_comparison}
    \resizebox{0.93\linewidth}{!}{
        \begin{tabular}{|c|c|c|c|c|c|c|c|c|c|}
            \hline
            \multirow{2}{*}{Method}   & \multirow{2}{*}{\begin{tabular}[c]{@{}c@{}}Tree Density\\(tree/m$^2$)\end{tabular}} & \multicolumn{4}{c|}{Latency (ms)}                   & \multicolumn{2}{c|}{Safety (m)} & \multirow{2}{*}{\begin{tabular}[c]{@{}c@{}}Bump Height\\(m)\end{tabular}} & \multirow{2}{*}{\begin{tabular}[c]{@{}c@{}}Traj. Length\\(m)\end{tabular}}                                                                               \\
            \cline{3-8}
                                      &                                                                                     & TTA                                                 & Pathfinding                     & Controller                                                                & Total                                                                      & Avg                    & Min         &                             &        \\
            \hline
            \multirow{3}{*}{GP-Nav}   & 0                                                                                   & \multirow{3}{*}{700.18}                             & \multirow{3}{*}{1961.53}        & \multirow{3}{*}{0.25}                                                     & \multirow{3}{*}{2661.95}                                                   & /                      & /           & \textcolor{mygreen}{13.31}  & 194.28 \\
            \cline{2-2}\cline{7-10}
                                      & 1/75                                                                                &                                                     &                                 &                                                                           &                                                                            & 3.94                   & 0.15        & \textcolor{red}{12.69}      & 194.25 \\
            \cline{2-2}\cline{7-10}
                                      & 1/18                                                                                &                                                     &                                 &                                                                           &                                                                            & 1.99                   & 0.03 [Fail] & \textcolor{blue}{12.56}     & 194.17 \\
            \hline
            \multirow{3}{*}{Proposed} & 0                                                                                   & \multicolumn{2}{c|}{\multirow{3}{*}{\textbf{7.26}}} & \multirow{3}{*}{\textbf{4.33}}  & \multirow{3}{*}{\textbf{11.58}}                                           & /                                                                          & /                      & 13.62       & \textcolor{mygreen}{180.60}          \\
            \cline{2-2}\cline{7-10}
                                      & 1/75                                                                                & \multicolumn{2}{c|}{}                               &                                 &                                                                           & \textcolor{red}{5.07}                                                      & \textcolor{red}{0.80}  & 15.25       & \textcolor{red}{180.66}              \\
            \cline{2-2}\cline{7-10}
                                      & 1/18                                                                                & \multicolumn{2}{c|}{}                               &                                 &                                                                           & \textcolor{blue}{3.07}                                                     & \textcolor{blue}{0.59} & 13.81       & \textcolor{blue}{183.24}             \\
            \hline
        \end{tabular}
    }
\end{table*}

\begin{figure}
    \centering
    \subfloat[Experiment without trees.]{\includegraphics[width=\linewidth]{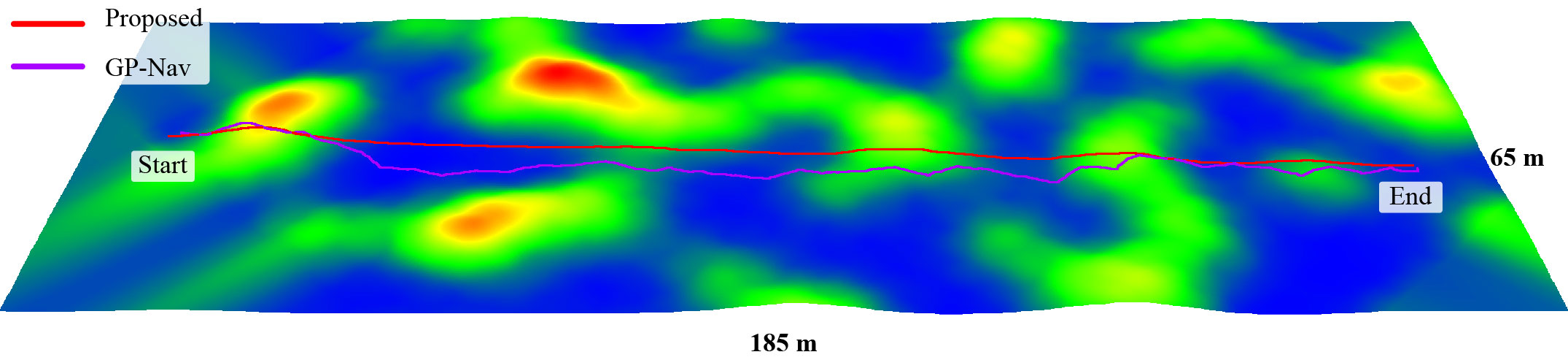}%
        \label{fig:simulated_comparison:no_tree}%
    } \\
    \subfloat[Experiment with the tree density of 1/75 tree/m$^2$.]{\includegraphics[width=\linewidth]{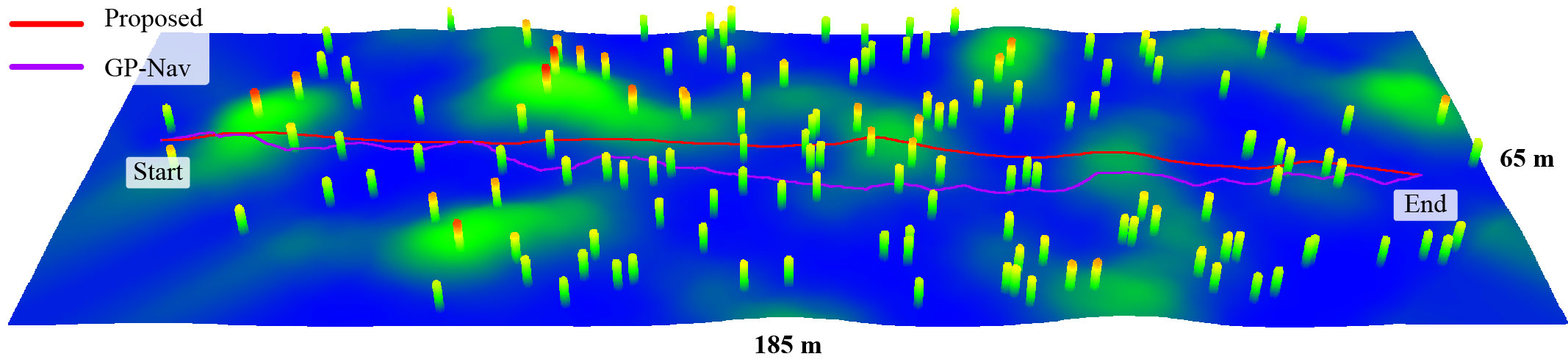}%
        \label{fig:simulated_comparison:sparse}%
    } \\
    \subfloat[Experiment with the tree density of 1/18 tree/m$^2$.]{\includegraphics[width=\linewidth]{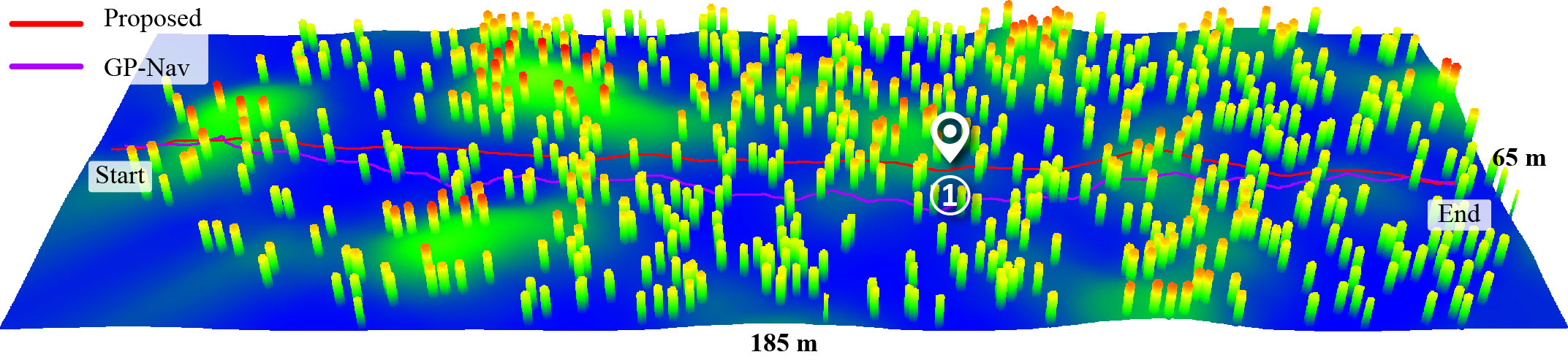}%
        \label{fig:simulated_comparison:dense}%
    }
    \caption{
        Trajectory comparisons in the simulated off-road forest environment with different tree densities.
        The red curve is the trajectory of the proposed method, and the purple curve is the trajectory of GP-Nav.
    }
    \label{fig:simulated_comparison}
\end{figure}

This section compares the proposed approach with a recent SOTA off-road navigation method: GP-Nav \cite{gp-navigation} as the baseline.
The comparisons are conducted in a randomized simulated forest generated by the YOPO-Sim with massive trees and rugged terrain, as shown in \cref{fig:simulation_environment}, where the depth image, LiDAR, and odometry are available.
Note that the proposed method takes the depth image and odometry as input, while GP-Nav takes the LiDAR point cloud and odometry as input.
GP-Nav is fine-tuned to ensure a fair comparison for better performance in the simulated environment.

The trees with a diameter of 0.5 m and a density of 0, 1/75, and 1/18 tree/m$^2$ are placed randomly in the environment.
The average and maximum slopes of the terrain are about $6.2^\circ$ and $25.2^\circ$, respectively.
The algorithms are evaluated in the same environment with the same start and goal positions \text{180 m} apart and the same average speed of 1 m/s.
The average performance statistics are shown in \cref{tab:simulated_comparison,fig:simulated_comparison}, and the snapshot of the proposed method with the tree density of 1/18 tree/m$^2$ is shown in \cref{fig:experiment:snapshots:sparse}.
The better calculation time is bolded, and the better metric of each density is highlighted in green, red, and blue, respectively.

Firstly, the computational time is compared.
Traditional methods like GP-Nav include TTA, pathfinding, and a controller.
To accelerate TTA, GP-Nav skips the mapping stage and computes the cost map directly from the raw LiDAR point cloud through a sparse Gaussian process regression on the GPU.
Its pathfinding is based on the RRT* algorithm, which requires a lot of samples to find the feasible path, and the controller is based on the PID algorithm.
In contrast, the proposed method fuses TTA and pathfinding into a single neural network to obtain multiple candidate trajectories with only one forward inference, reducing the planning time from \text{2661 ms} to \text{7.26 ms}.
The controller of the proposed method is based on the MPC algorithm, which produces much smoother trajectories than the PID algorithm of GP-Nav.
In addition, the GP-Nav triggers replanning by distance to reduce the computation resource, while the proposed method triggers replanning by time to ensure safety.

Subsequently, the planning performance is compared, including the safety metric (the average and minimum distance to the nearest obstacle), the bump height (the absolute value of height change along the trajectory), and the trajectory length.
As illustrated, the proposed method achieves a better safety metric than GP-Nav.
And for the tree density of \text{1/18 tree/m$^2$}, GP-Nav fails to find a collision-free trajectory.
This is because the proposed method utilizes an explicit collision cost to avoid collisions with trees, while GP-Nav only implicitly considers it in the slope and step height costs.
For the bump height metric, GP-Nav performs better than the proposed method, as the proposed method does not consider long-term planning.
In addition, the long-range and high-precision LiDAR sensor can provide more accurate terrain information than the depth camera.
Regarding trajectory length, the proposed method achieves a shorter trajectory than GP-Nav.
This is because the proposed method prioritizes the goal position in the cost function, while GP-Nav prefers to explore flatter terrain.

\subsection{Real-World Experiment}

In this section, the proposed learning-based single-stage planner is evaluated in real-world using the \text{DIABLO} robot with a size of $54 \times 37 \times 49 \text{ cm}^3$ and a weight of $22.9 \text{ kg}$, as shown in \cref{fig:diablo_robot}.
The computational resource is limited to an NVIDIA Jetson Orin NX with 8 CPU cores and 1024 CUDA cores, which only takes about 25 ms for inference and 20 ms for the MPC solver.
The RealSense T265 camera is used for odometry, and the OAK-D-Pro stereo camera is adopted for depth image with an FOV of $80^\circ \times 55^\circ$ and a range of around 12 m.
Invalid pixels of the depth image are inpainted with the Navier-Stokes based method, and the resolution is reduced to $160 \times 32$ with the nearest-neighbor method.

\begin{figure}
    \centering
    \subfloat[Environment A with the average speed of 1.13 m/s.]{
        \begin{minipage}{0.95\linewidth}
            \includegraphics[width=\linewidth]{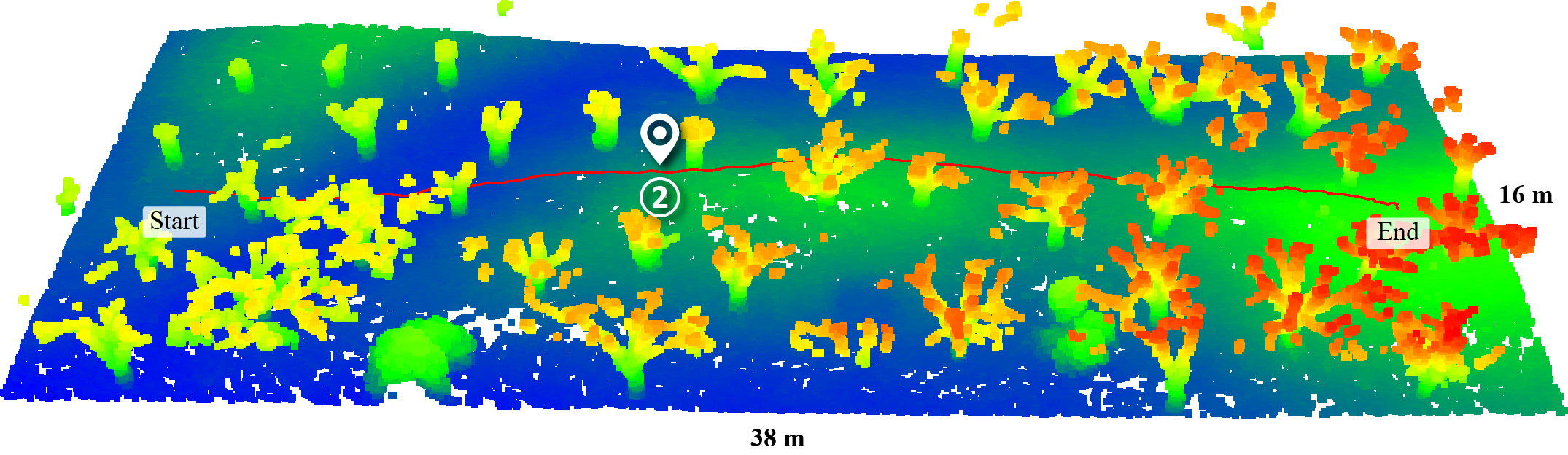}
            \includegraphics[width=\linewidth]{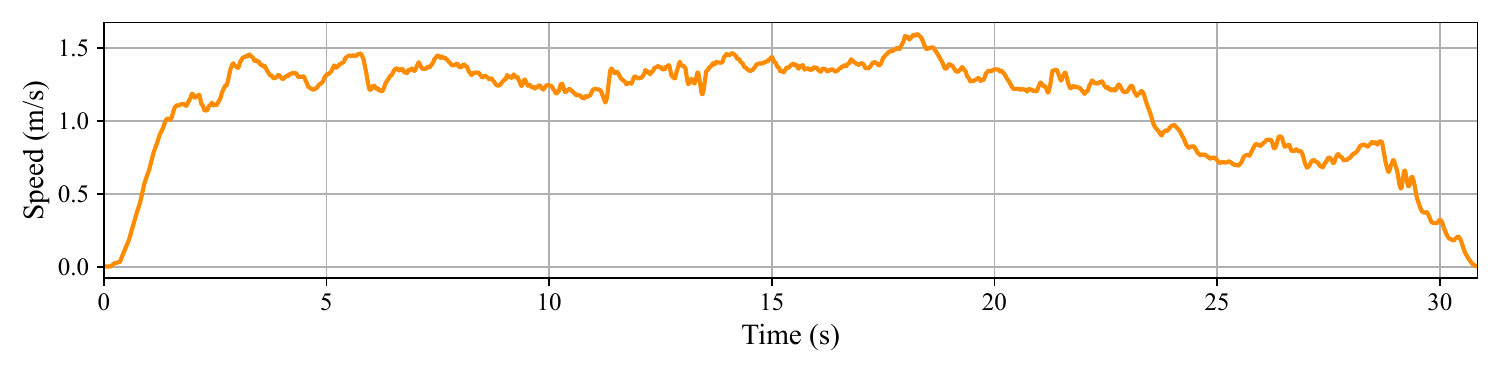}
        \end{minipage}%
        \label{fig:real_world_experiment:bag11}
    } \\
    \subfloat[Environment B with the average speed of 1.21 m/s.]{
        \begin{minipage}{0.95\linewidth}
            \includegraphics[width=\linewidth]{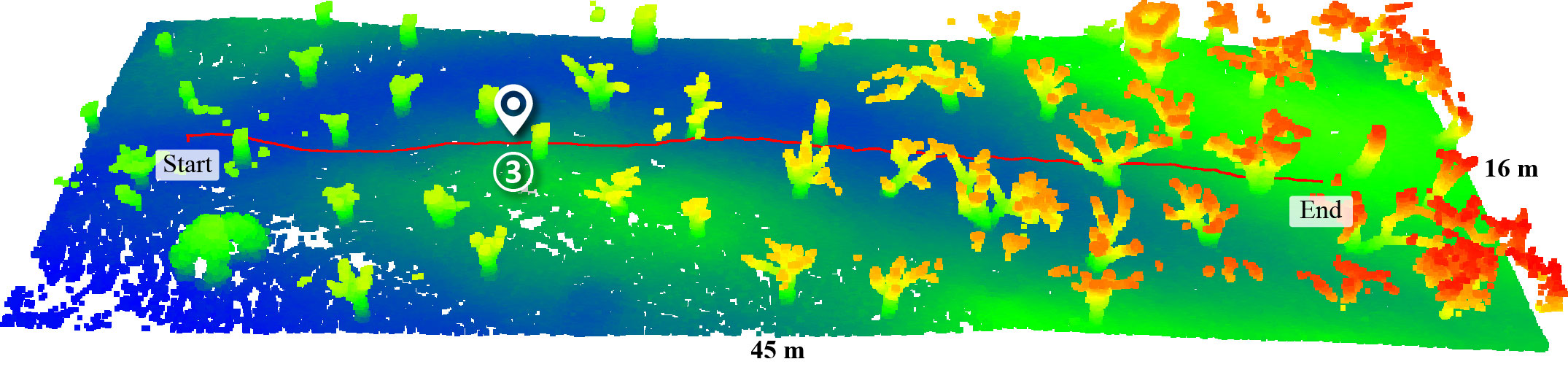}
            \includegraphics[width=\linewidth]{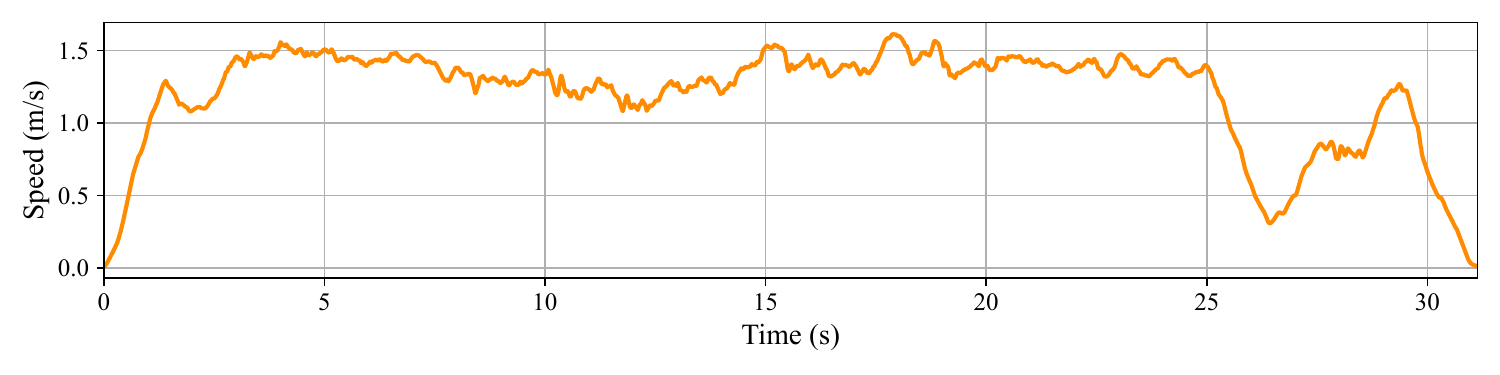}
        \end{minipage}%
        \label{fig:real_world_experiment:bag9}
    }
    \caption{
        Trajectories visualization of real-world experiments in forest terrain with velocity profiles.
    }
    \label{fig:real_world_experiment}
\end{figure}

As shown in \cref{fig:experiment:snapshots:bag9,fig:experiment:snapshots:bag11}, the real-world experiments are conducted in the off-road environment with massive trees and rugged terrain, which is challenging in terms of complex terrain, noisy depth images, limited sensing range, and computational resources.
The diameter of the trees is around \text{0.2 m}, and the tree density is about 1/12 tree/m$^2$.
The average and maximum slopes of the terrain are about $5.1^\circ$ and $18.6^\circ$, respectively.
The trajectories and velocity profiles of the robot are shown in \cref{fig:real_world_experiment}, and the metrics are summarized in \cref{tab:real-world_experiment}.
The robot executes the planned trajectories through the forest and achieves a maximum speed of 1.6 m/s, with an average safety distance of 1.35 m and a minimum of 0.4 m.
Note that the point cloud map is only for visualization and not for planning.
For details of the real-world experiment, please refer to the video.

\begin{table}
    \renewcommand\arraystretch{1.2}
    \centering
    \caption{Real-World Experiment}
    \label{tab:real-world_experiment}
    \resizebox{0.95\linewidth}{!}{
        \begin{tabular}{|c|c|c|c|c|c|c|c|}
            \hline
            \multirow{2}{*}{Env} & \multicolumn{3}{c|}{Latency (ms)} & \multicolumn{2}{c|}{Safety (m)} & \multirow{2}{*}{\begin{tabular}[c]{@{}c@{}}Bump Height\\(m)\end{tabular}} & \multirow{2}{*}{\begin{tabular}[c]{@{}c@{}}Traj. Length\\(m)\end{tabular}}                       \\
            \cline{2-6}
                                 & NN                                & Local                           & Total                                                                     & Avg                                                                        & Min  &      &       \\
            \hline
            A                    & \multirow{2}{*}{24.5}             & \multirow{2}{*}{20.4}           & \multirow{2}{*}{44.9}                                                     & 1.45                                                                       & 0.40 & 3.38 & 35.49 \\
            \cline{1-1}\cline{5-8}
            B                    &                                   &                                 &                                                                           & 1.23                                                                       & 0.52 & 3.25 & 38.49 \\
            \hline
        \end{tabular}
    }
\end{table}

\section{Conclusions}

In this letter, we propose an end-to-end off-road navigation framework, consisting of a high-performance, multi-sensor supported off-road simulator YOPO-Sim, a zero-shot transfer sim-to-real planner YOPO-Rally, and an MPC controller.
The proposed simulator shows competitive performance with mainstream simulators.
The planner integrates the TTA and the pathfinding into a single neural network trained by behavior cloning, where all the training data is generated in the simulator.
It demonstrates competitive performance in computation time and trajectory quality compared to the SOTA off-road planner.
Due to the slight discrepancy between the simulated and real-world depth images, the planner can be directly applied to the real-world without any fine-tuning.
Finally, the planner is extensively evaluated in the simulator and real-world forest environments.
Our future work will focus on integrating more sensors into the simulator and improving the performance of the planner.

\bibliographystyle{IEEEtran}
\bibliography{paper}

\end{document}